\documentclass[a4paper,fleqn]{cas-dc}

\usepackage[numbers]{natbib}
\usepackage{subfigure}
\usepackage{algorithm}
\usepackage{algorithmic}
\usepackage{amsmath,amsfonts,amsthm}
\usepackage{paralist}
\usepackage{multirow,multicol}
\usepackage{tabularx}
\usepackage{booktabs}
\usepackage{subfigure}
\usepackage{colortbl}
\usepackage{makecell}
\usepackage{color}

\def\tsc#1{\csdef{#1}{\textsc{\lowercase{#1}}\xspace}}
\tsc{WGM}
\tsc{QE}
\tsc{EP}
\tsc{PMS}
\tsc{BEC}
\tsc{DE}

\begin{document}
\let\WriteBookmarks\relax
\def\floatpagepagefraction{1}
\def\textpagefraction{.001}
\shorttitle{FADO Network}
\shortauthors{Wei Peng et~al.}

\title [mode = title]{FADO: \underline{F}eedback-\underline{A}ware \underline{D}ouble C\underline{O}ntrolling Network for Emotional Support Conversation}                      
\tnotemark[1]

\tnotetext[1]{This document is the results of the research
project funded by the National Natural Science
Foundation of China (No.U21B2009).}

\author[1,2]{Wei Peng}[ orcid=0000-0001-8179-1577]
\cormark[2]
\ead{pengwei@iie.ac.cn}

\address[1]{Institute of Information Engineering, Chinese Academy of Sciences,  Beijing, China}

\author[3]{Ziyuan Qin}[style=chinese]
\cormark[2]
\ead{qandy1992@gmail.com}

\author[1,2]{Yue Hu}
\cormark[1]
\ead{huyue@iie.ac.cn}

\author[1,2]{Yuqiang Xie}
\ead{xieyuqiang@iie.ac.cn}

\author[1,2]{Yunpeng Li}
\ead{liyunpeng@iie.ac.cn}

\address[2]{School of Cyber Security, University of Chinese Academy of Sciences, Beijing, China}

\address[3]{West China Biomedical Big Data Center of West China Hospital of Sichuan University, China}

\cortext[cor1]{Corresponding author}
\cortext[cor2]{Equal Contribution}

\begin{abstract}
Emotional Support Conversation (ESConv) aims to reduce help-seekers' emotional distress with a supportive strategy and response. It is essential for the supporter to select an appropriate strategy according to the feedback of the help-seeker (e.g., emotion change during dialog turns, etc) in ESConv. However, previous methods mainly rely on the dialog history to select the strategy and ignore the help-seeker's feedback, causing wrong and user-irrelevant strategy predictions. Meanwhile, these methods only model the context-to-strategy flow but pay less attention to the strategy-to-context flow involving the strategy-related context for generating strategy-constrained responses. In this paper, a \textbf{F}eedback-\textbf{A}ware \textbf{D}ouble C\textbf{O}ntrolling Network (\textbf{FADO}) is proposed to make a strategy schedule and generate supportive responses. The core modules in \textbf{FADO} include a dual-level feedback strategy selector and a double control reader, where the former leverages the turn-level and conversation-level feedback to encourage or penalize strategies, and the latter constructs a novel strategy-to-context flow to generate strategy-constrain responses. Besides, a strategy dictionary is designed to enrich the semantic information of the strategy and improve the quality of the strategy-constrained response. Experimental results on ESConv indicate that the proposed FADO achieves state-of-the-art performance in terms of strategy selection and response generation. Our code is available at https://github.com/Thedatababbler/FADO.
\end{abstract}


\begin{highlights}
\item An method is proposed to make a strategy schedule and generate a strategy-constrained response in emotional support conversations.
\item The dual-level feedback strategy selector incorporates the dual-level feedback of the help-seeker at the turn level and conversation level.
\item The double control reader effectively constructs a context-to-strategy flow and a strategy-to-context flow between the strategy and dialog history.
\item The proposed method achieves state-of-the-art performance in terms of strategy selection and response generation.
\end{highlights}


\maketitle

\section{Introduction}
\begin{figure}[t]
\centering
\includegraphics[width=0.5\textwidth]{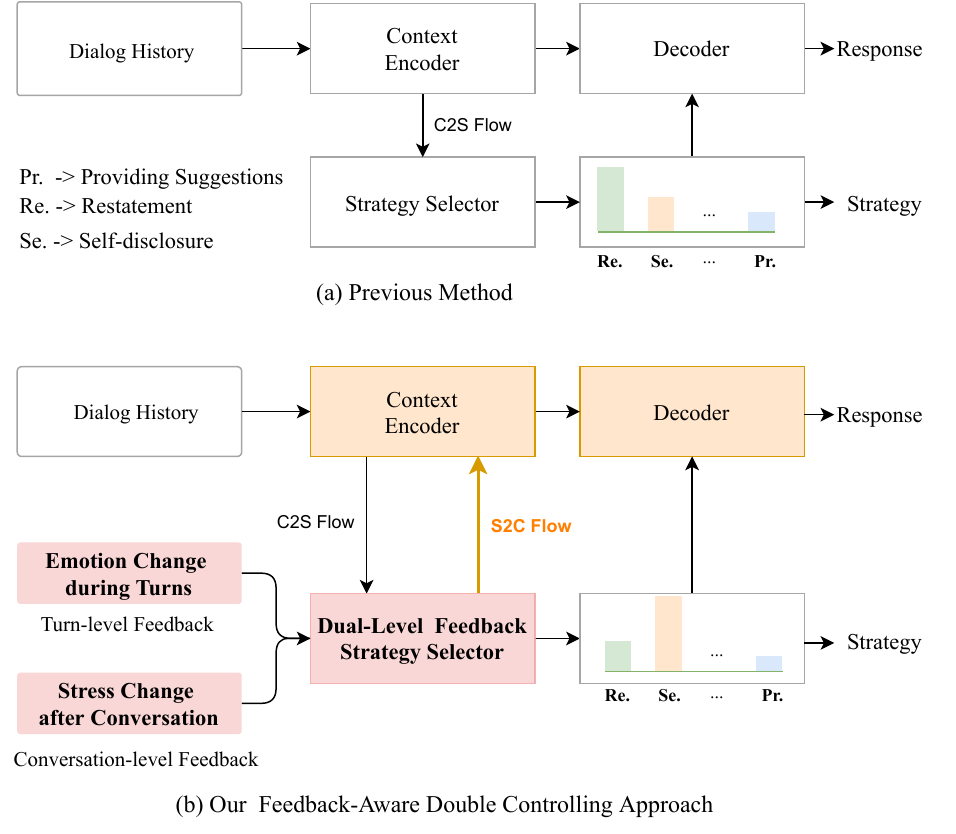}
\caption{An example to compare previous work (a) that only utilizes the dialog history and our work (b) that considers the dual-level user's feedback (in {\color{red}red}), such as the emotion change and stress change, and S2C flow (in \textcolor[rgb]{0.98,0.75,0.14}{orange}) for supportive response generation. C2S and S2C represent context-to-strategy and strategy-to-context, respectively.
}
\label{fig:exam}
\end{figure}
In recent years, the research on dialog systems is booming in the field of natural language processing (NLP), and a variety of neural models {\cite{DBLP:conf/aaai/SerbanSBCP16,DBLP:conf/aaai/XingWWHZ18,DBLP:conf/aaai/GaoBLLS19,DBLP:conf/aaai/Sun0XYX20,chen2021gog,Ni2022HiTKGTG,DBLP:conf/coling/WangCLF0022}} have been proposed, which can generate fluent and coherent responses. To reduce the gap of high-level and complex ability between dialog systems and human beings, researchers are trying to complete a new and {challenging} {emotional support conversation} task \cite{DBLP:conf/acl/LiuZDSLYJH20} that focuses on reducing individuals' emotional distress with an appropriate strategy and providing supportive responses to a help-seeker \cite{Rains2020SupportSE,heaney2008social,DBLP:conf/chi/SlovakGF15}.
In this case, it is important to train an emotional support dialog system with the proper strategy \cite{Zwaan2012ABD,DBLP:journals/coling/ZhouGLS20} so that it can be applied to a wide range of scenarios, including mental health support (comforting a depressed help-seeker and providing suggestions)
, social interactions (encouraging the help-seeker), etc. \cite{DBLP:conf/acl/LiuZDSLYJH20}. 

As for the emotional support scenery, studies \cite{hill1999helping,DBLP:conf/acl/LiuZDSLYJH20,DBLP:conf/acl/TuLC0W022} have shown that it is significant for dialog agents to select an appropriate supportive strategy that can guide response generation and comfort the help-seeker's emotion. 
However, there are some issues with strategy selection and supportive response generation. Firstly, as shown in Fig. \ref{fig:exam} (a), most studies \cite{DBLP:conf/acl/LiuZDSLYJH20,DBLP:conf/acl/TuLC0W022} simply leverage the dialog history to select a strategy and optimize it with the ground truth. {{For example, the work \cite{DBLP:conf/acl/TuLC0W022} predicts the strategy distribution based on the encoded context representation}}. However, it ignores the dual-level feedback of the help-seeker (as shown in \textcolor[rgb]{0.98,0,0}{red} in Fig. \ref{fig:exam} (b)), {{thus leading}} to wrong and user-irrelevant strategy prediction (inability to understand the user's feelings).
{Specifically, the dialog system should give priority to the strategies that consider help-seekers' positive feedback} \cite{scott2000understanding}. {{This paper focuses on two aspects of feedback: turn-level feedback (the user's feeling during turns) and conversation-level feedback (the user's feeling after the conversation).}}
Secondly, current methods only integrate the context-to-strategy flow but ignore the strategy-to-context flow (as shown in \textcolor[rgb]{0.98,0.75,0.14}{orange} in Fig. \ref{fig:exam} (b)) { in the encoding phase}. 
Specifically, the system not only needs to consider the influence of the dialog history on the strategy but also relies on the representation of the strategy to focus on the strategy-related context { in the encoding phase} to generate strategy-constrained responses.
Therefore, this paper mainly addresses two challenges: 1) leveraging the dual-level feedback of the help-seeker to schedule the strategy, 
and 2) modeling the strategy-to-context flow.

To this end, this paper proposes a novel \textbf{F}eedback-\textbf{A}ware \textbf{D}ouble C\textbf{O}ntrolling Network (\textbf{FADO}) for emotional support conversation. The core modules include a \textbf{D}ual-level \textbf{F}eedback \textbf{S}trategy Selector \textbf{(DFS)} and a \textbf{D}ouble \textbf{C}ontrol \textbf{R}eader \textbf{(DCR)}. Specifically, \textbf{DFS} designs turn-level feedback (locally reflects the current users' feelings) and conversation-level feedback (globally represents the users' global states) to encourage or penalize strategies {during the strategy selection process}. Secondly, \textbf{DCR} develops a context-to-strategy flow and a strategy-to-context flow to focus on the strategy-related dialog context and generate strategy-constrained responses. Besides, \textbf{FADO} introduces a strategy dictionary that records the description of strategies to enrich their semantic information and improve the quality of the strategy-constrained response.

The contributions of this paper are summarized as follows:
\begin{itemize}
\item A feedback-aware double controlling network is proposed to make a strategy schedule and generate a strategy-constrained response in emotional support conversations.
\item To make an accurate and user-relevant strategy selection, the \textbf{DFS} incorporates the feedback of the help-seeker at the turn level and conversation level.
\item To focus on the strategy-related context and generate strategy-constrained responses, the \textbf{DCR} effectively constructs context-to-strategy and strategy-to-context flows between the strategy and dialog history.
\item Experiments on the dataset indicate that the \textbf{FADO} achieves the State-Of-The-Art {(SOTA)} performance in terms of strategy selection and response generation.
\end{itemize}

\section{Related Work}

\subsection{Emotional Conversation Model}
A critical research topic for dialog systems is to design agents with emotional intelligence \cite{picard2003affective}, which is defined as the ability to perceive, integrate, understand, and regulate emotions \cite{salovey1997emotional}.
Recently, studies on emotional chatting have grown rapidly \cite{DBLP:conf/aaai/ZhouHZZL18,DBLP:conf/acl/SongZLXH19,DBLP:conf/coling/ZhangCXX20,DBLP:journals/tois/HuangZG20,DBLP:journals/corr/abs-2203-12254}. One notable work \cite{DBLP:conf/aaai/ZhouHZZL18} is the Emotional Chatting Machine (ECM), which generates appropriate emotional responses conditioning on a pre-specified label with a memory network \cite{DBLP:conf/emnlp/MillerFDKBW16}. {The work \cite{Wen2022DynamicIM} focuses on dynamic interactions during the information fusion process and proposes a dynamic interactive multi-view memory network model to integrate interaction information for recognizing emotions.} In \cite{DBLP:conf/coling/ZhangCXX20}, a novel knowledge-aware incremental transformer with multi-task learning is designed to leverage commonsense knowledge and encode multi-turn contextual utterances for improving emotion recognition performance. The work \cite{DBLP:conf/acl/SongZLXH19} introduces an emotional dialog system (EmoDS) that can generate responses with a coherent structure for a post and express the desired emotion explicitly or implicitly within a unified framework.
\begin{figure*}
\centering
\includegraphics[width=0.99\textwidth]{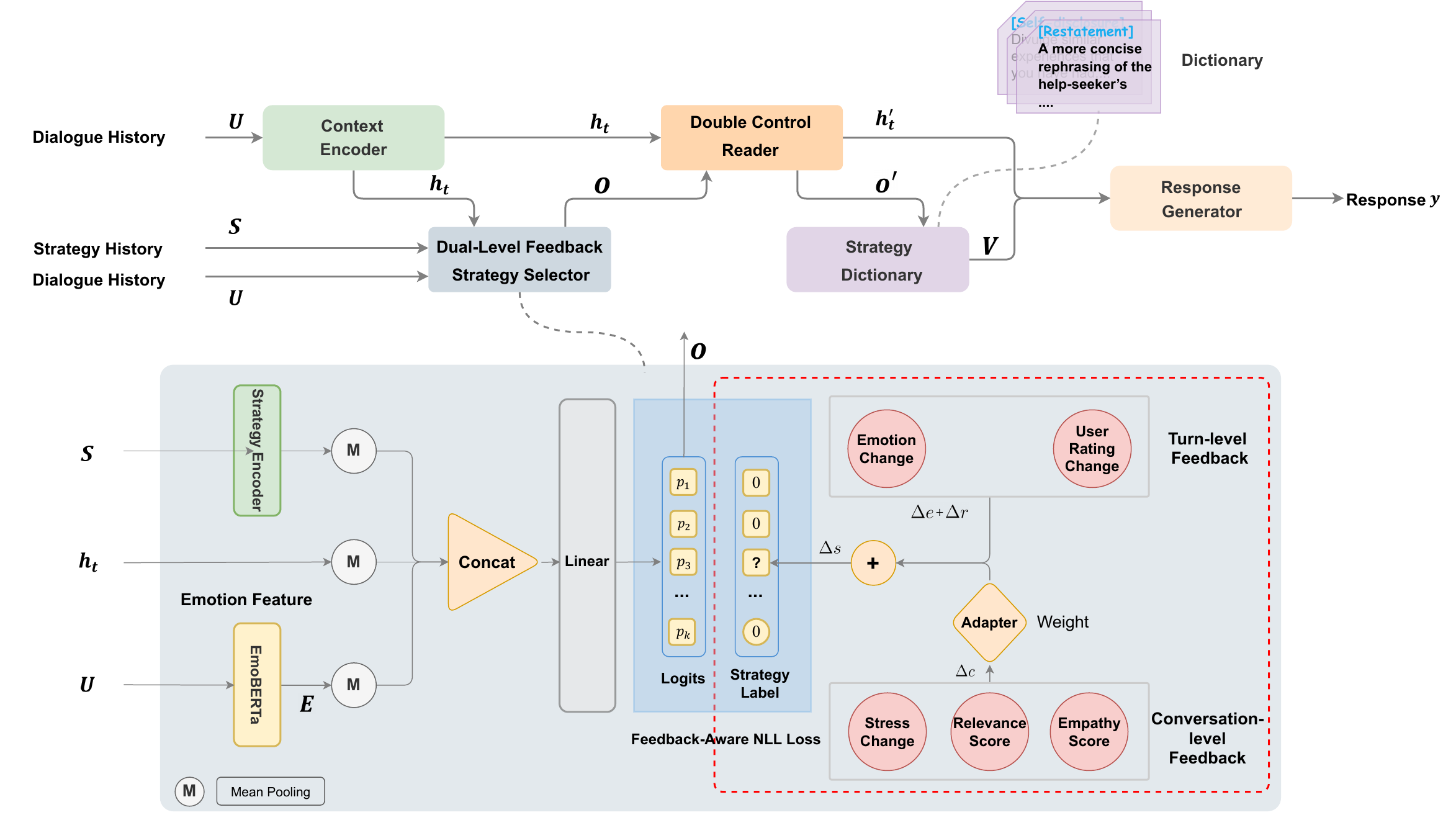}
\caption{The structure of our framework, which consists of a context encoder, a dual-level feedback strategy selector, a double-control reader, a strategy dictionary, and a response generator. The gray dotted lines indicate the detail of the corresponding module, and the red dotted line indicates that the part only exists in the training process. }
\label{fig:model}
\end{figure*}
\subsection{Empathetic Conversation Model}
{According to the survey \cite{DBLP:journals/inffus/MaNXC20}}, different from the task of emotional chatting, where the agent generates emotional responses with the given emotion, the task of empathetic dialog generation requires the model to understand the feelings of other agents and then respond appropriately \cite{DBLP:conf/acl/RashkinSLB19,DBLP:conf/emnlp/LinMSXF19,DBLP:conf/emnlp/MajumderHPLGGMP20,DBLP:conf/acl/ZhengLCLH21,DBLP:journals/corr/abs-2210-03884}. For instance, the literature \cite{DBLP:conf/emnlp/LinMSXF19} proposes an empathetic generation strategy that relies on two key ideas for empathetic dialog generation, including emotion grouping and emotion mimicry. The literature \cite{DBLP:conf/emnlp/MajumderHPLGGMP20} leverages commonsense knowledge
to capture the situation of the user and uses this additional information to improve the ability of empathy expression to output responses. In \cite{DBLP:conf/acl/ZhengLCLH21}, a multi-factor hierarchical framework is developed to model the communication mechanism, dialog act, and emotion of empathy expression. In {\cite{Wang2021EmpatheticRG}, a novel graph-based model with multi-hop reasoning is developed to model the emotional causality of the empathetic conversation.} {The work \cite{Lee2022ExploringTR} investigates the repetition problem in the empathetic dialog generative model.} However, different from previous tasks, an emotional support conversation task needs to explore the users' problems and reduce their emotional distress with an appropriate strategy and supportive response generation \cite{DBLP:conf/acl/LiuZDSLYJH20}.

\subsection{Emotional Support Conversation Model}
Recently, emotional support conversation has received much attention \cite{DBLP:conf/acl/LiuZDSLYJH20,DBLP:conf/acl/TuLC0W022,DBLP:conf/ijcai/00080XXSL22,DBLP:journals/corr/abs-2202-13047}. Generally, emotional support conversation models can be divided into two categories: explicitly modeling psychological factors at the cognitive level with the graph network \cite{DBLP:conf/ijcai/00080XXSL22} and utilizing the strategy \cite{DBLP:conf/acl/LiuZDSLYJH20,DBLP:conf/acl/TuLC0W022} for generating supportive responses. For instance, the literature \cite{DBLP:conf/ijcai/00080XXSL22} {proposes} a global-to-local hierarchical graph network to capture the psychological features and model hierarchical relationships between them to generate a supportive response. As for {strategy}-aware generation, the literature \cite{DBLP:conf/acl/LiuZDSLYJH20} {uses} a special token to represent each strategy and {appends} the strategy token before the response to make a generation. In \cite{DBLP:conf/acl/TuLC0W022}, a mixed strategy-aware model integrating a pre-trained commonsense language model COMET \cite{DBLP:conf/acl/BosselutRSMCC19} is developed to respond skillfully. However, these studies ignore dual-level feedback and simply optimize the strategy with the ground truth, thus failing to encourage/penalize strategies that obtain {positive}/negative feedback. Furthermore, the strategy-to-context flow is less considered in the above methods during the emotional support conversation.

\section{Problem Formulation}
\label{sec:3}
Before introducing our approach, the problem formulation of the task is provided below. Given a multi-turn emotional support dialog that consists of $M$ utterances $\boldsymbol{U} = (\boldsymbol{u}_1^{(i)}, \boldsymbol{u}_2^{(i)}, \dots, \boldsymbol{u}_M^{(i)})$ in the dialog history and the target response $\boldsymbol{y}$, where $i$ indicates the $i$-th conversation. For brevity, the superscript $(i)$ is omitted in the remaining content. To effectively utilize the information of the strategy chain for strategy selection, in addition to the contextual token feature $\boldsymbol{H} = (\boldsymbol{h_1}, \boldsymbol{h_2}, \dots, \boldsymbol{h_T})$, 
the historical strategy category feature $\boldsymbol{S} = (\boldsymbol{s_1}, \boldsymbol{s_2}, \dots, \boldsymbol{s_J})$ is introduced, where $T$ and $J$ denote the number of dialog tokens and turns in the dialog history, respectively.
Our model aims to predict a strategy and generate supportive responses with the given $\boldsymbol{U}$ and $\boldsymbol{S}$.

\section{Approach}

The structure of the proposed framework is presented in Fig. \ref{fig:model}, which consists of the context encoder, {DFS}, {DCR}, the strategy dictionary, and the response generator. Firstly, the context encoder obtains the contextual representation. Then, {DFS} takes the strategy-relevant feature as the input and incorporates the help-seeker's feedback to encourage or penalize strategies for strategy selection. Subsequently, {DCR} makes double control flows (i.e., context-to-strategy flow and strategy-to-context flow) between the strategy and the dialog history to generate a strategy-constrained response. Afterward, the strategy dictionary introduces the description to enrich the strategy information. Finally, the response generator outputs the supportive response.

\subsection{Context Encoder}
The context encoder ${\mathtt{Enc_{cxt}}}$ based on BlenderBot \cite{Roller2021RecipesFB} {(following the previous work \cite{DBLP:conf/acl/TuLC0W022,DBLP:conf/ijcai/00080XXSL22})}, aims to obtain the representation of the dialog history, where BlenderBot is an open-domain dialog agent pre-trained on large-scale conversation corpora. 
Following the paper \cite{DBLP:conf/acl/LiuZDSLYJH20,DBLP:conf/ijcai/00080XXSL22}, each utterance is separated with the $\mathtt{[SEP]}$ token, and the $\mathtt{[CLS]}$ is the start token. The contextual representation $\boldsymbol{H} = (\boldsymbol{h_1}, \dots, \boldsymbol{h_t}, \dots, \boldsymbol{h_T})$ is modeled as follows:
\begin{equation}
\boldsymbol{H} = {\mathtt{Enc_{cxt}}}(\mathtt{[CLS]}, \boldsymbol{u}_1,\mathtt{[SEP]}, \boldsymbol{u}_2, \dots,\boldsymbol{u}_{M})
\label{eq:context}
\end{equation}
where $M$ and $T$ are the number of utterances and tokens in the dialog history, respectively; $\boldsymbol{h_t} \!\in\! \mathbb{R}^{d}$, $d$ is the dimension of the hidden state.

\subsection{Dual-level Feedback Strategy Selector}
In this section, the process of strategy selection by DFS is described first, and then the dual-level feedback is introduced.

\noindent
\textbf{Strategy Selection}
{DFS} takes the historical strategy category feature $\boldsymbol{S}$ and contextual representation $\boldsymbol{H}$ as inputs. Considering that the emotion of the help-seeker is an important factor for supporters to select a strategy, the emotional representation $\boldsymbol{E}$ is introduced and encoded by EmoBERTa \cite{DBLP:journals/corr/abs-2108-12009}, a pre-trained emotion-aware language model in conversation (the details are given in Section \ref{detailsofE}).

Specifically, the representation of historical strategy feature $\boldsymbol{s}$ is calculated by $\mathtt{Enc_{stra}}$ that has the same architecture as $\mathtt{Enc_{ctx}}$ with the mean-pooling operation:
\begin{eqnarray}\label{equ:1}
\begin{aligned}
\boldsymbol{s} &= {\mathtt {Mean}\verb|-|\mathtt{pooling}}({\mathtt{Enc_{stra}}}(\boldsymbol{S}))
\end{aligned}
\end{eqnarray}
where $\boldsymbol{s} \!\in\! \mathbb{R}^{d}$, $d$ is the dimension of the hidden state.

Similarly, the representations of context $\boldsymbol{c}$ and emotion $\boldsymbol{r}$ can be formulated as:
\begin{eqnarray}\label{equ:2}
\begin{aligned}
&\boldsymbol{c} = {\mathtt {Mean}\verb|-|\mathtt{pooling}}(\boldsymbol{h}_1, \boldsymbol{h}_2, \dots, \boldsymbol{h}_T) \\
&\boldsymbol{r} = {\mathtt {Mean}\verb|-|\mathtt{pooling}}(\boldsymbol{e}_1, \boldsymbol{e}_2, \dots, \boldsymbol{e}_T)
\end{aligned}
\end{eqnarray}
where $\boldsymbol{c} \!\in\! \mathbb{R}^{d}$, $\boldsymbol{r} \!\in\! \mathbb{R}^{d}$, the emotional representation $\boldsymbol{E} = (\boldsymbol{e}_1, \boldsymbol{e}_2, \dots, \boldsymbol{e}_T)$ is introduced and encoded by EmoBERTa as $\boldsymbol{E} = {\mathtt{EmoBERTa}}(\mathtt{[CLS]}, \boldsymbol{u}_1,\mathtt{[SEP]}, \boldsymbol{u}_2, \dots,\boldsymbol{u}_{M})$.

Finally, the distribution of the strategy $\boldsymbol{o}$ is defined as:
\begin{equation}
\boldsymbol{o} = \mathtt{MLP}(\mathtt{tanh}(\boldsymbol{W_o}^T[\boldsymbol{s};\boldsymbol{c};\boldsymbol{r}]+\boldsymbol{b_o}))
\label{eq:st}
\end{equation}
where $\boldsymbol{W_o} \!\in\! \mathbb{R}^{3d \times d}$, $\boldsymbol{b_o} \!\in\! \mathbb{R}^{d}$, $\boldsymbol{o} \!\in\! \mathbb{R}^{l}$, $l$ is the number of strategy categories, including seven categories (e.g., self-discourse, restatement) and one other category, {$\mathtt{MLP}$ represents multi-layer perceptron, $\mathtt{tanh}$ is an activation function.}

\noindent
\textbf{Dual-level Feedback}
The feedback of the help-seekers helps the system to understand their feelings to make a strategy selection. To incorporate the global and local states of the help-seeker simultaneously, DFS introduces turn-level and conversation-level feedback. The former reflects the current users' feelings, which can locally measure the benefit of strategies; the latter represents the users' global states, which can make guidance to correct offsets from a global perspective. 

Specifically, the \textbf{turn-level feedback} contains emotion and user rating change. Help-seekers express their emotion (to obtain the emotion score of positive polarity by the softmax function using the [CLS] representation of EmoBERTa) in each turn. 
Meanwhile, they give a user rating (in the dataset) after every two utterances from the supporter to score the effectiveness of the supporter on a 5-star scale. Based on these two factors, strategies can be locally measured by calculating the change in emotion $\Delta e$ and user rating $\Delta r$. As for \textbf{conversation-level feedback}, after each conversation, the help-seeker is required to rate their emotion and the effectiveness of the supporter according to the five-point Likert scale \cite{DBLP:conf/acl/LiuZDSLYJH20}. Here, three types of ratings are considered \footnote{1) the help-seeker's emotional stress after the conversation, 2) the relevance of the supporter's responses to the topic, and 3) the supporter's understanding and empathy of the help-seeker's feelings.} (Note: the sum of the total scores is denoted as $\Delta c$, {which} exists in the dataset). 

Finally, DFS introduces an adapter (i.e., the weight $\mu$) to {integrate} the two types of semantic information of turn-level and conversation-level feedback (the details of the calculation and statistical distribution on feedback are provided in Section \ref{dis-feedback}).
The dual-level feedback score $\Delta s$ utilized to encourage or penalize strategies in the calculation of the loss function can be formulated as:
\begin{equation}
\Delta s = \Delta e + \Delta r + \mu \Delta c
\label{eq:score}
\end{equation}
where $\mu$ is a hype-parameter; $\Delta s > 0$ indicates positive feedback; otherwise, it indicates negative feedback.

\begin{figure}[tbp]
\centering
\includegraphics[width=0.4\textwidth]{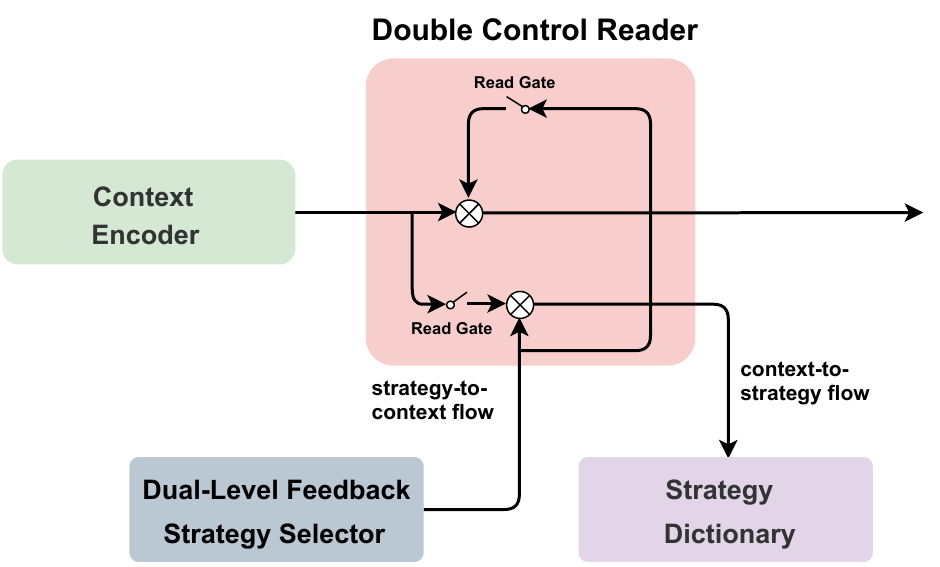}
\caption{The details of our double-control reader. }
\label{fig:model-2}
\end{figure}

\subsection{Double Control Reader}
To make a sufficient interaction between the strategy and the dialog history, the \textbf{DCR} develops a context-to-strategy flow and a strategy-to-context flow. Specifically, the context-to-strategy flow indicates that the model can leverage the context information to select a context-related strategy; the strategy-to-context flow indicates that the model can focus on the strategy-related context {in the encoding phase} and thus generate the strategy-constrained response. 

Motivated by \cite{DBLP:conf/aaai/ZhouHZZL18}, the read gates are adopted to control the information of flows:
\begin{eqnarray}\label{equ:double}
\begin{aligned}
\boldsymbol{g^c} &= {\mathtt {sigmoid}}(\boldsymbol{W_c}^T \boldsymbol{c} + \boldsymbol{b_c})  \\
\boldsymbol{g^o} &= {\mathtt {sigmoid}}(\boldsymbol{W_o}^T \boldsymbol{o} + \boldsymbol{b_o}) 
\end{aligned}
\end{eqnarray}
where $\boldsymbol{W_c} \!\in\! \mathbb{R}^{d \times l}$, $\boldsymbol{b_c} \!\in\! \mathbb{R}^{l}$, $\boldsymbol{g^c} \!\in\! \mathbb{R}^{l}$, $\boldsymbol{W_o} \!\in\! \mathbb{R}^{l \times d}$, $\boldsymbol{b_o} \!\in\! \mathbb{R}^{d}$, $\boldsymbol{g^o} \!\in\! \mathbb{R}^{d}$. The definitions of $\boldsymbol{c}$ and $\boldsymbol{o}$ are the same as those in Equations \ref{equ:2} and \ref{eq:st}. { $\mathtt{sigmoid}$ is an activation function.}

The read gates are utilized to read information from the context or strategy. Then, to make a trade-off between the original information and the updated information, the residual connection is introduced. The above process is described as:
\begin{eqnarray}\label{equ:dou}
\begin{aligned}
\boldsymbol{o'} = (1-\beta) \cdot \boldsymbol {o} &+ \beta  \cdot  \boldsymbol{g^c} \otimes \boldsymbol{o} \\
\boldsymbol{h_t'} = (1-\alpha) \cdot \boldsymbol {h_t} &+ \alpha  \cdot  \boldsymbol{g^o} \otimes \boldsymbol{h_t}
\end{aligned}
\end{eqnarray}
where $\boldsymbol{o'} \!\in\! \mathbb{R}^{l}$, $\boldsymbol{h'_t} \!\in\! \mathbb{R}^{d}$, $\alpha$ and $\beta$ are two hyper-parameters for controlling the weight.

\subsection{Strategy Dictionary}
To enrich the semantic information of the strategy and improve the quality of the strategy-constrained response, \textbf{FADO} introduces a strategy dictionary, where the key is a strategy and the value is the corresponding description. Different from the previous work \cite{DBLP:conf/acl/LiuZDSLYJH20} which generates the response conditioned on the strategy token, our work is based on the description for a deeper understanding of the strategy. Given the predicted strategy, following the attention mechanism in \cite{DBLP:conf/acl/TuLC0W022}, the description $\boldsymbol{V}$ is encoded and regarded as the input of the response generator (more details of the dictionary are provided in Section \ref{appendix:Definition of Strategies}).

\subsection{Response Generator}
The response generator generates a response $\boldsymbol{y}$ by combining the updated context representation $\boldsymbol{h_t'}$ and the description $\boldsymbol{V}$. Based on the BlenderBot \cite{Roller2021RecipesFB}, the conditional probability distribution is represented as: 
\begin{equation}\label{equ:decoder}
\boldsymbol{p} \left( y_z | \boldsymbol{y}_{< z}, \boldsymbol{h_t'}, \boldsymbol{V} \right) = \mathtt{Generator}(\boldsymbol{W}_{y < z}, \boldsymbol{h_t'}, \boldsymbol{V})
\end{equation}
where $z$ indicates the decoding time-step, $\boldsymbol{W}_{y < z}$ denotes the embedding of the generated tokens.

\subsection{Joint Training}
The loss function contains the feedback-aware negative log-likelihood (NLL) loss and the cross-entropy loss. 

As for strategy selection, considering the introduction of double control flows, the feedback-aware NLL loss between the updated strategy distribution $\boldsymbol{o'}$ and the true label based on the final feedback score $\Delta s$ is calculated. Specifically, the predicted strategy should be as close as possible to the ground truth if $\Delta s \ge 0$ (positive feedback); otherwise, the probability of the predicted strategy corresponding to the ground truth should be close to $0$. To realize the above process, the feedback-aware NLL loss is calculated as:
\begin{equation}
\label{Eq:gamma}
\mathcal{L}_{1}=
\begin{cases} \quad -\boldsymbol{ \hat{o}}\log( \mathtt{softmax}(\boldsymbol{o'}))  &if \quad {\Delta s > 0};\\
-\boldsymbol{ \hat{o}} \log( 1- \mathtt{softmax}(\boldsymbol{o'})) &if \quad {\Delta s \le 0};
\end{cases}
\end{equation}
where $\boldsymbol{\hat{o}}$ is the true strategy label.

As for response generation, the standard cross-entropy loss of the response generation is optimized as:
\begin{equation} \label{equ:l2}
\mathcal{L}_{2}=-\sum_{z=1}^{Z} \log \boldsymbol{p} \left( y_z | \boldsymbol{y}_{< z}, \boldsymbol{h_t'}, \boldsymbol{V} \right)
\end{equation}

The final joint objective is obtained as $\mathcal{L}=\mathcal{L}_{1}+\mathcal{L}_{2}$.
\begin{table}[t]
\centering
\begin{tabular}{lrrr}
	\toprule
	\textbf{Category}                         & \textbf{Total}           & \textbf{Supporter}   & \textbf{Seeker}     \\ \midrule
	\# dialogues                      & 1,053           & - & -    \\
	Avg. Minutes per Chat           & 22.6            & - & -    \\
	\# Workers               & 854             & 425         & 532        \\
	\# Utterances                   & 31,410          & 14,855      & 16,555     \\
	Avg. length of dialogues    & 29.8            & 14.1        & 15.7       \\
	Avg. length of utterances    & 17.8            & 20.2        & 15.7    \\\bottomrule
\end{tabular}
\caption{Statistics of ESConv.}
\label{tab:stats}
\end{table}

\section{Experiment}
\subsection{Dataset}
Our method and the comparison methods are evaluated on
the Emotional Support Conversation dataset, ESConv \cite{DBLP:conf/acl/LiuZDSLYJH20}, which is collected by crowdworkers in a help-seeker and supporter mode. The dataset contains relatively long conversations (29.8 utterances on average) with 31,410 utterances. The partition of train/dev/test subsets in the official dataset is adopted \cite{DBLP:conf/acl/LiuZDSLYJH20}. Help-seekers will give their user rating after every two utterances from the supporter, and the emotional score in this paper will be obtained by EmoBERTa \cite{DBLP:journals/corr/abs-2108-12009}. The overall statistics of the 1,053 ESConv examples are presented in Table \ref{tab:stats}.

\begin{table*}
\centering
\resizebox{0.97\linewidth}{!}{
	\begin{tabular}{lcccccccc}
		\toprule
		\textbf{Model} &\textbf{ACC}(\%)\,$\uparrow$ & \textbf{PPL}\,$\downarrow$ & \textbf{R-L}\,$\uparrow$ & \textbf{B-2}\,$\uparrow$ &\textbf{B-3}\,$\uparrow$ &\textbf{B-4}\,$\uparrow$& \textbf{D-1}\,$\uparrow$&
		\textbf{D-2}\,$\uparrow$\\ 
		\midrule
		Multi-Task Transformer \cite{DBLP:conf/acl/RashkinSLB19}	&-& 65.31                & 14.26     & 4.82  & 1.79 & 1.16  & 1.89  & 10.72    \\
		MoEL \cite{DBLP:conf/emnlp/LinMSXF19} &-	& 62.93                & 14.21   & 5.02  & 1.90 & 1.14  & 2.71  & 14.92  \\
		MIME \cite{DBLP:conf/emnlp/MajumderHPLGGMP20}   &- & 43.27               & 14.83   & 4.82  & 1.79 & 1.03  & 2.56  & 12.33   \\
		\midrule
		DialoGPT  \cite{DBLP:conf/acl/LiuZDSLYJH20}	   & -            &{15.51}   & 15.26  & 5.13  & - & -  & -  & -      \\
		GLHG \cite{DBLP:conf/ijcai/00080XXSL22}	&- & {15.67}   & {16.37}  & {7.57}  & {3.74} & {2.13}  &{3.50}  & {21.61}   \\
		BlenderBot-Joint$^*$ \cite{DBLP:conf/acl/LiuZDSLYJH20} &28.11	& 16.11   & 15.57  & 5.83  & 2.31 & 1.55  & 2.60  & 19.15  \\
		MISC$^*$ \cite{DBLP:conf/acl/TuLC0W022}	& 31.48 & {16.70}   & {16.74}  & {7.22}  & {3.29} & {2.06}  &{3.71}  & {20.98}   \\
		\textbf{FADO (Ours)	}&\textbf{32.90} & {15.72}   & \textbf{17.53}  & \textbf{8.00}  & \textbf{4.00} & \textbf{2.32}  &\textbf{3.84}  & \textbf{21.84}   \\
		\midrule
		MISC$^\diamond$ \cite{DBLP:conf/acl/TuLC0W022}	& 31.63 & {16.16}   & {17.91}  & {7.31}  & {-} & {2.20}  &\textbf{4.41}  & {19.71}   \\
		\textbf{FADO$^\diamond$ (Ours)	}&\textbf{32.41} & \textbf{15.52}   & \textbf{18.09}  & \textbf{8.31}  & \textbf{4.36} & \textbf{2.66}  &{3.80}  & \textbf{21.39}   \\
		\bottomrule
\end{tabular}}
\caption{Automatic evaluation results on ESConv. $^*$ indicates that the performance is reproduced. Other results are obtained from the paper \cite{DBLP:conf/ijcai/00080XXSL22}. $^\diamond$ indicates that the results are based on the re-split dataset in MISC for a fair comparison. } 
\label{tab:main-exp}
\end{table*}

\subsection{Experimental Setting}
\label{setting}
The BlenderBot-small \cite{Roller2021RecipesFB} is employed as the context encoder and response generator like papers \cite{DBLP:conf/acl/TuLC0W022,DBLP:conf/ijcai/00080XXSL22}, and the Pytorch framework \cite{Paszke2017AutomaticDI} is used. EmoBERTa \cite{DBLP:journals/corr/abs-2108-12009} is exploited to calculate the emotional scores and representations during each turn. The epoch is set to $3$ with a learning rate of $3e$-$5$ and a linear warmup of $100$ steps. The batch size of training is set to $16$. The AdamW \cite{Loshchilov2017FixingWD} optimizer is used for training with $\beta_1 = 0.9$, $\beta_2 = 0.999$, and $\epsilon = 1e$-$8$. Following \cite{DBLP:conf/acl/LiuZDSLYJH20}, this study adopts the 
Top-$p$ sampling with $p=0.9$, temperature $\tau=0.7$, and a repetition penalty of $1.0$. The hyper-parameters $\alpha$ and $\beta$ in the DCR are set to $0.2$ and $0.2$ according to the grid search, respectively. Besides, $\mu$ is set to $0.5$. The experiments are implemented on Tesla V100-16G GPU. The
source code \footnote{https://github.com/Thedatababbler/FADO.} will be released to facilitate future research work. {For data preprocessing, this study directly utilizes the data in the official benchmark \footnote{https://github.com/thu-coai/Emotional-Support-Conversation}. As for the implementation of baselines, the source code of MISC \footnote{https://github.com/morecry/MISC} has been released so that researchers can reproduce results with the provided official dataset.}

\subsection{Details of EmoBERT}
\label{detailsofE}
EmoBERTa \cite{DBLP:journals/corr/abs-2108-12009}, a pre-trained language model based on {RoBERTa}, is designed for the Emotion Recognition in Conversation (ERC) task. It achieves the SOTA performance on several well-known public datasets (e.g., MELD, IEMOCAP, etc.) and is simple to migrate to other downstream tasks. Therefore, this study adopts an off-the-shelf detector \footnote{https://github.com/tae898/erc.}, EmoBERTa-base, as our feature extractor to obtain the emotional score and emotional representation of help-seekers. The emotional score is obtained by the softmax function using the [CLS] representation of EmoBERTa-base. The experimental results indicate that emotional states are beneficial for both strategy selection and response generation.

\subsection{Description of Strategy Dictionary}
\label{appendix:Definition of Strategies}
To enrich the semantic information of the strategy and improve the quality of the strategy-constrained response, \textbf{FADO} introduces a strategy dictionary whose key is a strategy and value is the corresponding description. Following the literature \cite{DBLP:conf/acl/LiuZDSLYJH20}, the description of each strategy is defined as follows.\\
\noindent\textbf{Question} Asking for information related to the
problem to help the help-seekers articulate the issues that they face. Open-ended questions are best, and closed questions can be used to obtain specific information.\\
\noindent\textbf{Restatement or Paraphrasing} A simple, more concise \\ rephrasing of the help-seekers' statements that could help them see their situation more clearly.\\
\noindent\textbf{Reflection of Feelings} Articulate and describe the help-\\seekers' feelings.\\
\noindent\textbf{Self-disclosure} Divulge similar experiences that you have had or emotions that you share with the help-seeker to express your empathy.\\
\noindent\textbf{Affirmation and Reassurance} Affirm the help-seekers' \\ strengths, motivation, and capabilities, and provide reassurance and encouragement.\\
\noindent\textbf{Providing Suggestions} Provide suggestions about how to change the situation, but be careful to not overstep and tell them what to do.\\
\noindent\textbf{Information} Provide useful information to the help-seeker with data, facts, opinions, resources, etc., or by answering questions.\\
\noindent\textbf{Others} Exchange pleasantries and use other support strategies that do not fall into the above categories.

\subsection{Baselines}
The following baselines containing empathetic and {pre-trained} dialog models are taken for comparison: 

\begin{itemize}
\item \textbf{Multi-Task Transformer} \cite{DBLP:conf/acl/RashkinSLB19}: A variation of the Transformer that has an additional task for predicting the emotion.
\item \textbf{MoEL} \cite{DBLP:conf/emnlp/LinMSXF19}: A Transformer-based model that combines representations from multiple decoders to improve the response empathy. 
\item \textbf{MIME} \cite{DBLP:conf/emnlp/MajumderHPLGGMP20}: Another {Transformer-based} model that leverages the emotional polarity and mimicry for empathetic generation. 
\item \textbf{BlenderBot-Joint} \cite{DBLP:conf/acl/LiuZDSLYJH20}: BlenderBot-Joint is an open-domain conversational agent {trained with multiple communication skills, including empathetic responding}, and it appends a special strategy token before the response utterances to make a generation. {The implementation details are presented at the link.} \footnote{https://github.com/thu-coai/Emotional-Support-Conversation}
\item \textbf{GLHG} \cite{DBLP:conf/ijcai/00080XXSL22}: The global-to-local hierarchical graph network proposed in this paper to capture the multi-source information and model hierarchical relationships for emotional support generation.
\item \textbf{MISC} \cite{DBLP:conf/acl/TuLC0W022}: {It} designs a mixed strategy-aware model to introduce COMET to capture user’s instant mental state and generate supportive responses with a mixed representation of strategies. {The implementation details are provided at the link.} \footnote{https://github.com/morecry/MISC}
\end{itemize}

\subsection{Evaluation Metrics}
\noindent
\textbf{Automatic Evaluations} (1) For the strategy selection, the prediction accuracy (ACC) of the strategy is taken as the evaluation metric. (2) For the response generation, the conventional PPL (perplexity), BLEU-$n$ (B-$n$) \cite{DBLP:conf/acl/PapineniRWZ02}, Distinct-$n$ (D-$n$) \cite{DBLP:conf/naacl/LiGBGD16} and ROUGE-L (R-L) \cite{lin2004rouge} are take as the evaluation metrics. BLEU-$n$ and ROUGE-L are widely used for evaluating the quality of language generation. BLEU \cite{DBLP:conf/acl/PapineniRWZ02} (bilingual evaluation understudy) is an algorithm for evaluating the quality of that text that is machine-translated from one natural language to another. Quality is considered to be the correspondence between a machine's output and that of a human. ROUGE \cite{lin2004rouge} is a set of metrics and a software package used for evaluating automatic summarization and machine translation in NLP. Distinct-$n$ measures the proportion of unique $n$-grams in the generated responses to evaluate generation diversity, and it is a reference-irrelevant metric.

\noindent
\textbf{Human A/B Evaluations} In previous studies, human evaluation is usually conducted by crowdsourcing workers who rate responses on a scale from 1 to 5 from the aspects of fluency, relevancy, etc. However, {the} criteria can vary widely between different individuals. Therefore, this study adopts the human A/B evaluations for a high inter-annotator agreement. Given two models A and B, like FADO and one baseline, three annotators are asked to choose the better response for each of the $150$ sub-sampled test instances. For objectivity, annotators include those with and without background knowledge (task-related). The final results are determined by majority voting. If the three annotators reach three different conclusions, the fourth annotator will be brought in. Following \cite{DBLP:conf/acl/LiuZDSLYJH20}, the aspects include Fluency (Flu.), Identification (Ide.), Comforting (Com.), Suggestion (Sug.) and Overall (Ove.). Specifically, (1) Fluency: which bot’s responses are more fluent and understandable? (2) Identification: which bot explores your situation more in-depth and is more helpful in identifying problems? (3)
Comforting: which bot is more skillful in comforting you? (4) Suggestion: which bot gives you more helpful suggestions for the problems? (5) Overall: which bot’s emotional support do you prefer?
\begin{table}[h]
\centering
\resizebox{\linewidth}{!}{
	\begin{tabular}{lccccc}
		\toprule
		\textbf{Comparisons} &
		\textbf{Aspects} &
		\textbf{Win} &
		\textbf{Lose} &
		\textbf{Tie} \\
		\midrule
		&  Flu. &\textbf{23.3$^\dag$}& 9.3& 67.4\\
		& Ide. &\textbf{65.3$^\ddag$}& 19.4& 15.3\\
		FADO vs. BlenderBot & Com.&\textbf{52.7$^\ddag$}& 26.0& 21.3\\
		& Sug.&\textbf{65.3$^\ddag$}&14.0&20.7\\
		& Ove.&\textbf{60.7$^\ddag$}&16.0&23.3\\
		\midrule
		&  Flu. &\textbf{22.7$^\text{ }\text{ }$} & 14.7& 62.6\\
		&  Ide. &\textbf{47.3$^\dag$}&30.7&22.0\\
		FADO vs. GLHG  & Com.&\textbf{42.7$^\text{ }\text{ }$}&34.7&22.6\\
		& Sug.&\textbf{54.0$^\ddag$}&29.3&16.7\\
		& Ove.&\textbf{48.0$^\dag$}&25.3&26.7\\
		\midrule
		&  Flu. &\textbf{20.0$^\text{ }\text{ }$} & 13.3& 66.7\\
		& Ide. &\textbf{43.3$^\text{ }\text{ }$}&36.0&20.7\\
		FADO vs. MISC & Com.&\textbf{51.3$^\ddag$}&29.3&19.4\\
		& Sug.&\textbf{48.7$^\text{ }\text{ }$}&36.7&14.6\\
		& Ove.&\textbf{44.0$^\dag$}&26.7&29.3\\
		\toprule
\end{tabular}}
\caption{
	Human A/B evaluation results. ${\dag}$ and ${\ddag}$ represent the improvement with $p$-value $< 0.1/0.05$, respectively; \textit{Tie} indicates that the responses from both models are equal.  
}
\label{table:h_results}
\end{table}
\section{Experimental Results}
\subsection{Automatic Evaluations}
The automatic results are shown in Table \ref{tab:main-exp}. Note that the {BlenderBot} and {MISC} are reproduced \footnote{https://github.com/morecry/MISC.} in this paper because the data is re-split in MISC, which is not the original benchmark dataset in the official ESConv. 
For the other baselines, the results are obtained from the paper \cite{DBLP:conf/ijcai/00080XXSL22} that keeps the train/test partition unchanged. For consistency, the results on the official ESConv are reported, and for a fair comparison, the results on the re-split dataset in MISC are also reported. As shown in Table \ref{tab:main-exp}, FADO shows a strong ability to predict a more accurate strategy, and it obtains the SOTA performance for response generation, showing the effectiveness of the proposed model.
Specifically, the higher accuracy of the strategy prediction indicates that the dual-level feedback is conducive to {making} an appropriate schedule {for the strategy selection task}. As for the supportive response generation task, FADO obtains the best results over the baseline models, indicating the higher quality of our generated responses. In summary, {FADO} achieves the SOTA performance in terms of both strategy selection and response generation.
\begin{table}[b]	
\centering
\begin{tabular}{lllll}
\toprule
&\textbf{ACC $\uparrow$}  & \textbf{B-2 $\uparrow$} & \textbf{D-1 $\uparrow$} & \textbf{R-L $\uparrow$} \\ \midrule
{FADO} 	&\textbf{32.90}& \textbf{8.00} & \textbf{3.84} 	& {17.53}
\\	\midrule
{w/o~Strategy History}	& {31.24}	& {7.52}  & {3.78} & {17.10}\\
{w/o~Emotion Feature} & {30.20}	& {7.62} & {3.33}  & {17.05}\\
{w/o~TL \& CL Feedback}	& {31.16}	& {7.53}  & {3.55}  & {17.19}\\
{w/o~TL Feedback} 	& {31.19}	& {7.67}  & {3.50}  & \textbf{17.60}\\	
{w/o~CL Feedback} 	& {30.96}	& {7.84}  & {3.66}  & {16.98}\\	
{w/o~S2C \& C2S Flow} 	& {30.39}	& {7.12}  & {3.10}  & {16.35}\\	
{w/o~S2C Flow} 	& {32.02}	& {7.38}  & {3.20}  & {16.77}\\	
{w/o~C2S Flow} 	& {32.12}	& {7.68}  & {3.60}  & {16.89}\\	
{w/o~Strategy Dictionary} 	& {-}	& {6.12}  & {3.75}  & {16.40}\\	
\toprule
\end{tabular}
\caption{\label{tab:ablation} The results of ablation study on each components. }
\end{table}
\subsection{Human A/B Evaluations}
As shown in Table \ref{table:h_results}, the human evaluations are consistent with the automatic evaluations. The baselines are all {Pre-trained Language Models (PLMs)} that help make a more competitive comparison. It can be seen that the responses from FADO are {much} more preferred than those of the baselines in terms of the five aspects. For example, compared with MISC, our model is superior in terms of the Com. metric, indicating that FADO can generate comforting responses with appropriate strategies based on the dual-level feedback of the help-seeker (e.g., emotion change and stress change). As for the Sug. metric, FADO obtains more noticeable advancements than GLHG, which demonstrates the importance of modeling the strategy flow. Besides, it is noted that FADO does not significantly outperform {PLMs} in the Flu. metric, and this is probably attributed to the powerful expressing ability of the {PLMs}, but FADO still achieves decent improvements. To sum up, our method improves the results in all aspects, highlighting the necessity in incorporating users' feedback and modeling double control flows between the strategy and dialog history.

\section{Analyses}
\subsection{Ablation Study}
To verify the effectiveness of the components in FADO, an ablation study is conducted, and the results are presented in Table \ref{tab:ablation}. It can be seen that each component is beneficial to the final result. (1) Both the strategy history and emotion feature contribute to the model's performance, especially in terms of the ACC metric, which indicates the importance of the strategy-relevant features for strategy selection. (2) To evaluate the \textbf{DFS}, Turn-Level (\textbf{TL}) feedback and Conversation-Level (\textbf{CL}) feedback are removed, both of which have an impact on the final results, demonstrating the effectiveness of \textbf{DFS}. Furthermore, the \textbf{CL} feedback makes a considerable contribution compared to the \textbf{TL} feedback, showing the greater necessity of global feedback information. { Removing the TL feedback causes a slight increase in R-L, and the possible reason is that the TL feedback indicates the local and immediate feedback, which affects the current dialog state and helps generate more distinct responses. Additionally, there is a trade-off between the Distinct metric and ROUGE metric, i.e., the higher Distinct metric, the lower ROUGE metric. Therefore, as shown in Table \ref{tab:ablation}, the TL feedback shows better performance in the D-1 metric, causing a slight increase in R-L.} (3) \textbf{DCR} that consists of the Context-to-Strategy (\textbf{C2S}) flow and the Strategy-to-Context (\textbf{S2C}) flow improves the performance, and removing these two flows leads to a significant drop in ACC (-2.51\%) and ROUGE-L (-1.18\%), which proves that such bi-directional flows are effective for modeling the interaction between the dialog history and strategy. (4) The strategy dictionary is also {important} to the performance of the proposed model.
\begin{figure*}
	\centering
	\subfigure[predicted by the FADO.]{
		\begin{minipage}[t]{0.33\linewidth}
			\centering
			\includegraphics[width=2in]{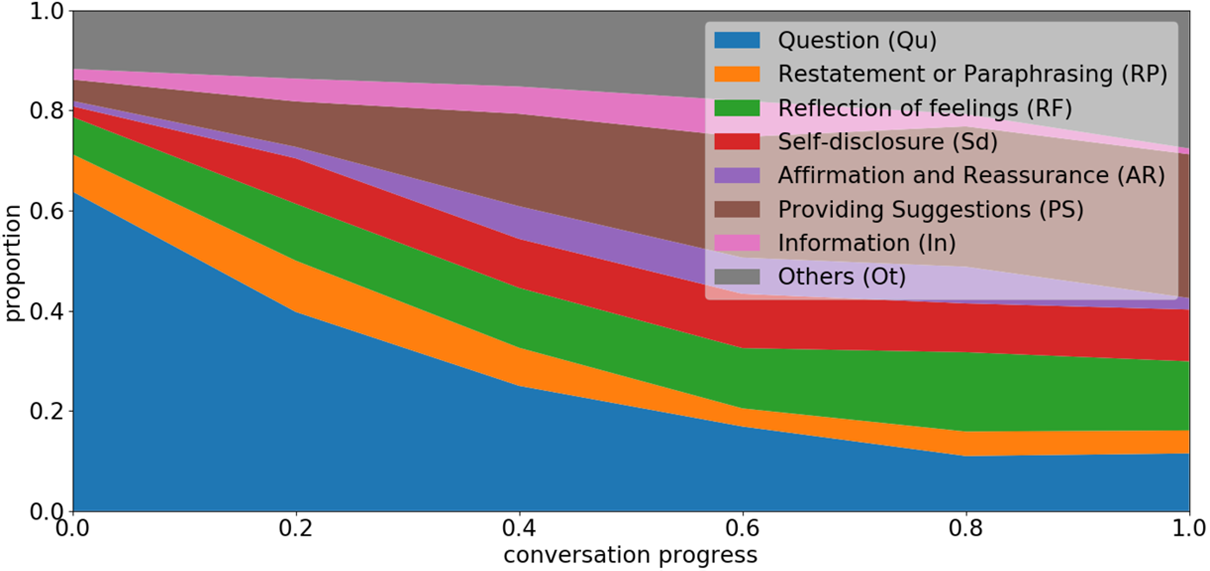}
		\end{minipage}%
	}%
	\subfigure[from the test set.]{
		\begin{minipage}[t]{0.33\linewidth}
			\centering
			\includegraphics[width=2in]{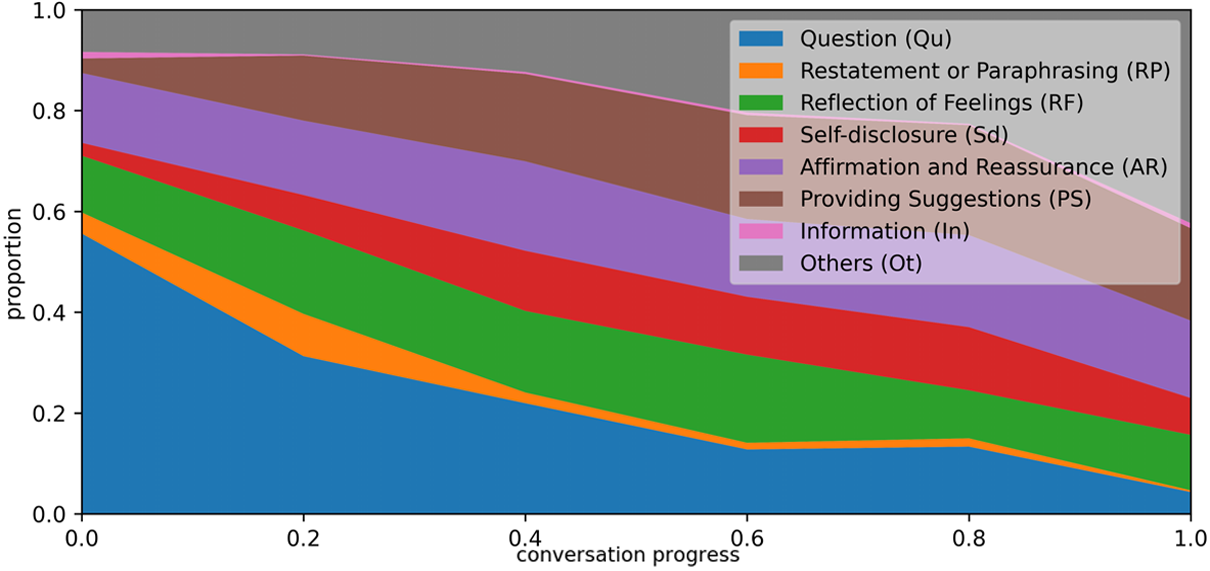}
		\end{minipage}
	}%
	\subfigure[predicted by the MISC.]{
		\begin{minipage}[t]{0.33\linewidth}
			\centering
			\includegraphics[width=2in]{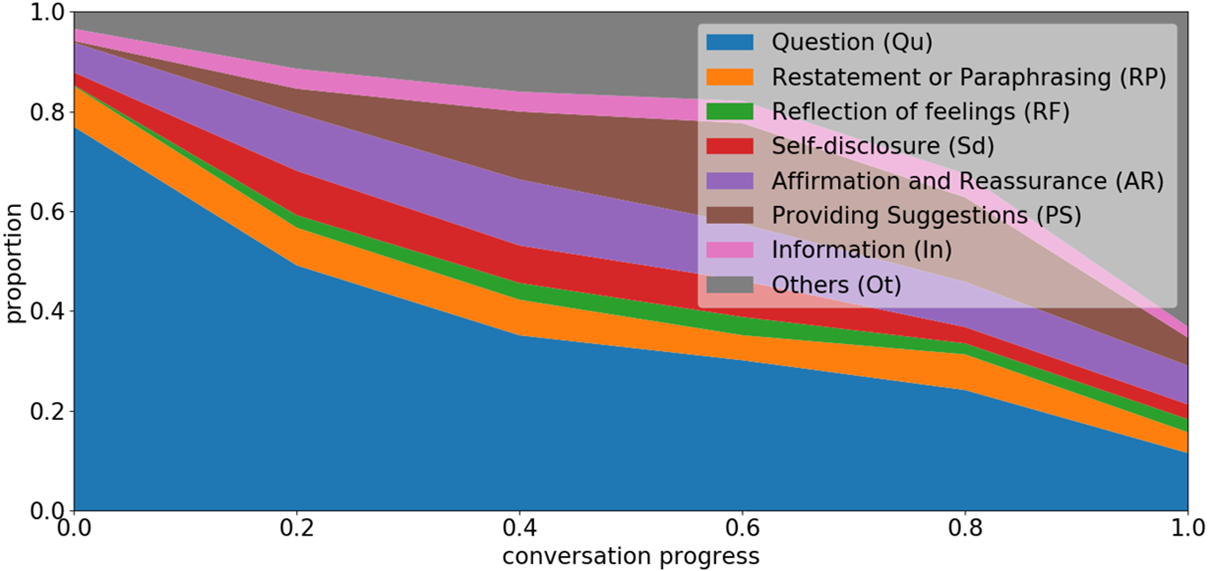}
		\end{minipage}%
	}%
	\caption{The strategy distribution in the different stage of conversation. }
	\label{distribution}
\end{figure*}

\begin{table}[h]
\centering
\begin{tabular}{lcccc}
\toprule
\textbf{Method}& \textbf{ACC}\,$\uparrow$ &\textbf{B-2}\,$\uparrow$ & \textbf{D-1}\,$\uparrow$&\textbf{R-L}\,$\uparrow$\\ 
\midrule
BlenderBot& 28.11&{5.83}&{2.60}&{15.57}\\
\midrule
FADO ($\alpha$=0)    & 32.12         & 7.68       & 3.60                      & {16.89}                     \\
\textbf{FADO ($\alpha$=0.2)} & \textbf{32.90} & \textbf{8.00} & \textbf{3.84}            & \textbf{17.53}           \\
FADO ($\alpha$=0.5) & 31.86          & 7.39       & 3.47                     & 17.05                    \\
FADO ($\alpha$=0.8) & 31.51          & 7.65       & 3.51                     & 17.16                    \\
FADO ($\alpha$=1.0)   & 31.03            & 7.52       & 3.74                     & 17.20  \\  
\toprule
\end{tabular}
\caption{Results of different settings on the S2C flow.}
\label{tab:xxx}
\end{table}

\begin{table}[h]
\centering
\begin{tabular}{lcccc}
\toprule
\textbf{Method}& \textbf{ACC}\,$\uparrow$ &\textbf{B-2}\,$\uparrow$ & \textbf{D-1}\,$\uparrow$&\textbf{R-L}\,$\uparrow$\\ 
\midrule
BlenderBot& 28.11&{5.83}&{2.60}&{15.57}\\
\midrule
FADO($\beta$=0)            & 31.16          & 7.39          & \textbf{3.88}           & 17.11          \\
\textbf{FADO($\beta$=0.2)} & \textbf{32.90} & \textbf{8.00} & {3.84} & \textbf{17.53} \\
FADO($\beta$=0.5)          & 32.63          & 7.43          & 3.70          & 17.13          \\
FADO($\beta$=0.8)          & 32.71          & 7.57          & 3.50          & 17.04          \\
FADO($\beta$=1.0)            & 31.61          & 7.79          & 3.78          & 17.25   \\
\toprule
\end{tabular}
\caption{\label{tab:beta} The results of different settings on the C2S flow.}
\end{table}

\subsection{Strategy Distribution Analysis}
The benchmark dataset ESConv suggests that emotional support usually follows a certain order of strategy flow. Specifically, the conversation progress can be divided into six intervals. Then, for all the testing conversations, the proportions of different strategies are counted and {drawn} on the six intervals at six points, and  the strategy distributions are depicted in Fig. \ref{distribution}. 
Our model has a very similar distribution to the ground-truth distribution, which indicates that FADO can mimic strategy learning like human supporters to provide more effective emotional support. Meanwhile, FADO obtains better results than MISC on \textit{Providing Suggestions} and \textit{Other} strategy, demonstrating the effective strategy modeling ability of our method.

\subsection{Strategy-to-Context Flow Analysis}
To investigate how the {S2C} flow affects the performance of the FADO, this study manually sets the hyper-parameter $\alpha$. (Note that: the {C2S} flow is considered in the previous work, and this result is presented in the latter Section \ref{c2s}.) As shown in Table \ref{tab:xxx},
the performance of the FADO achieves a peak on all the aspects after considering S2C flow ($\alpha=0.2$), which indicates that utilizing the S2C flow is necessary for the model's performance. \\

\subsection{Context-to-Strategy Flow Analysis}
\label{c2s}
The {C2S} flow has been considered in the previous work. To investigate how the C2S flow affects the performance, this paper manually sets the hyper-parameter $\beta$ based on the $\alpha=0.2$ setting. As listed in {Table \ref{tab:beta}}, the model achieves a decent improvement after considering the C2S flow ($\beta=0.2$), which indicates that utilizing the C2S flow is also beneficial for improving the model's performance.

\begin{table*}[htbp]
\centering	
\resizebox{\linewidth}{!}{
	\begin{tabular}{|l|l|}
	\hline
	\multicolumn{2}{|c|}{\textbf{Case One: Context}}\\
	\hline
	Help-seeker &\makecell[l]{Tired, I want to sleep all my problems away.}\\
	\hline
	Supporter &\makecell[l]{How long has this been going on?} \\
	\hline
	Help-seeker & \makecell[l]{Well the boyfriend for awhile the other thing for a day or two, I got a hold of a therapist \\ but they won't talk to me for two days.} \\
	\hline
	&\makecell[l]{......}\\
	\hline
	Help-seeker & \makecell[l]{I can't sleep through the day because \textbf{{\color{red}I have kids}}... so it's typically late afternoon to \\ sleep mostly.}\\
	\hline
	\multicolumn{2}{|c|}{\textbf{Response}}\\
	\hline
	{{Ground-Truth}} & \makecell[l]{\textbf{\color{blue}(\textit{Affirmation and Reassurance})} It sounds like you are doing the very best you can under \\ difficult circumstances. You should feel very proud about that.}\\
	\hline
	MISC & \makecell[l]{\textbf{\color{cyan}(\textit{Providing Suggestions})} I can understand why you feel helpless. Do you have \\ any family members? }\\
	\hline
	\textbf{FADO}  &\makecell[l]{\textbf{\color{blue}(\textit{Self-disclosure})} I understand. I also \textbf{{\color{red}have a daughter.}} She is very difficult to deal with. \\ Do you talk to your friends or family about that?}\\
	\hline
	
	\multicolumn{2}{|c|}{\textbf{Case Two: Context}}\\
	\hline
	Help-seeker &\makecell[l]{Hi.}\\
	\hline
	Supporter &\makecell[l]{Hi there, how are you feeling?} \\
	\hline
	&\makecell[l]{......}\\
	\hline
	Help-seeker &\makecell[l]{As a single mother it's pretty hard to go to search for food bank. I am spending most of  \\ my time in \textbf{\color{red}looking for a job.}}\\
	\hline
	Supporter &\makecell[l]{Are there any \textbf{\color{red}organisations} that you could just call and explain your predicament to?} \\
	\hline
	Help-seeker & \makecell[l]{\textbf{\color{red}I don't know any} as I never faced such situation before ... Now life is becoming \\ misearable.} \\
	\hline
	\multicolumn{2}{|c|}{\textbf{Response}}\\
	\hline
	{{Ground-Truth}} & \makecell[l]{\textbf{\color{blue}(\textit{Providing Suggestions})} Do you have friends who could help you, even with \\ child care?}\\
	\hline
	MISC & \makecell[l]{\textbf{\color{cyan}(\textit{Information})} Have you tried looking for a job? \textit{(Repetition)} }\\
	\hline
	\textbf{FADO}  &\makecell[l]{\textbf{\color{blue}(\textit{Providing Suggestions})} I truly understand that, maybe you could try to contact your \textbf{\color{red}local} \\ \textbf{\color{red}government} or \textbf{\color{red}local agencies.}}\\
	\hline
	\end{tabular}}
\caption{Case study of the generated responses by the baselines and FADO. Some words are omitted due to space limit. {\color{cyan}Cyan} indicates the inconsistency between the strategy and response, {\color{blue}blue} indicates consistency, and {\color{red}Red} represents the relevant information.}
\label{tab:gena} 
\end{table*}

\subsection{Case Study}
{Table \ref{tab:gena}} shows some examples from the FADO and MISC qualitatively. More cases for other baselines are provided in Section \ref{morecase}. For case one, our FADO generates an appropriate response by giving a similar experience \textit{I also have a daughter} to express empathy although it selects a different strategy \textit{Self-disclosure} from the ground truth. The selected strategy and the generated response are consistent. By contrast, MISC makes an inconsistent response (as shown in {\color{cyan}cyan}). In case two, MISC outputs the repetition \textit{looking for a job} that has appeared in the context. By contrast, FADO guides the emotional reply with \textit{I truly understand} and provides suggestions \textit{contact your local government}. Additionally, MISC usually {makes} an inconsistency between the strategy and response (as shown in {\color{cyan}cyan}). However, FADO produces more consistent responses (as shown in {\color{blue}blue}), indicating that the strategy-to-context flow can enable the model to generate consistent and strategy-constrained responses.

\begin{table}[h]
\centering
\resizebox{0.95\linewidth}{!}{
	\begin{tabular}{lcc}
	\toprule
	\textbf{Model}	& \textbf{Consistency Score} & $\Delta$  \\ \midrule
	~{BlenderBot-Joint} 	& {41.67}	&  -
	\\
	~{MISC} & {52.00}	& + 10.33	\\
	~{FADO}	& \textbf{57.33}	&	+ 15.66	\\
	\toprule
	\end{tabular}}
\caption{\label{tab:consistancy} The results of consistency analysis between strategy selection and response generation. }
\end{table}
\subsection{Consistency Analysis}
In previous work, the consistency evaluation between the predicted strategy and generated response is not considered, which leads to the contradiction between the strategy prediction task and response generation task. For instance, one baseline predicts the \textit{providing suggestions} strategy, but the response is \textit{why do you feel sad, have you talked to your friends?} (\textit{question} strategy). Therefore, this study introduces consistency analysis into the emotional support conversation. Specifically, three annotators are asked to determine whether the strategy corresponds to the response based on $150$ sub-sampled test instances. To the best of our knowledge, this is the first work to make a consistency analysis in the emotion support conversation task. As shown in Table \ref{tab:consistancy}, FADO achieves the SOTA performance in terms of the consistency score (gains 15.66\% score), which indicates that the proposed DCR can leverage the strategy-to-context flow to generate strategy-constrained responses.
\begin{figure*}
\centering
\includegraphics[width=\textwidth]{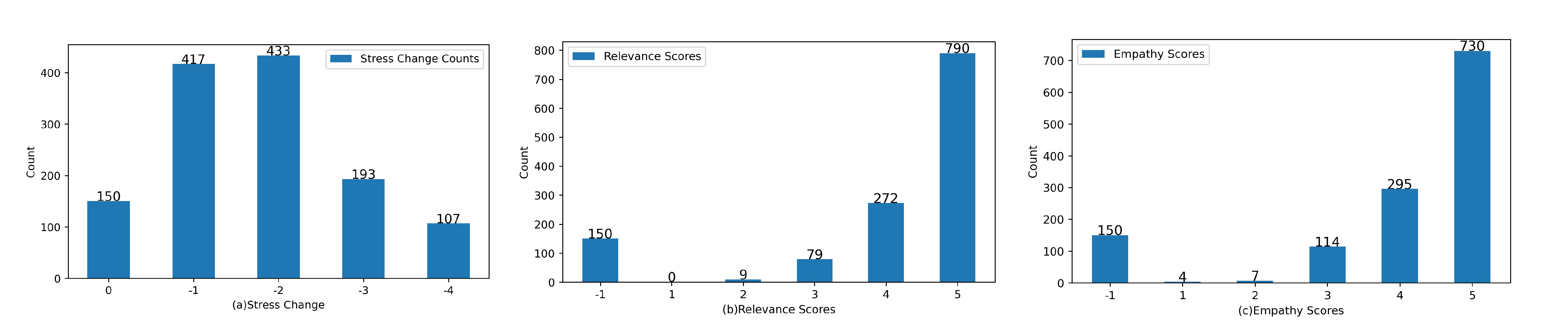}
\caption{Distribution of the feedback. (a) Stress Change, (b) Relevance Scores, and (c) Empathy Scores.
}
\label{fig:dis}
\end{figure*}

\begin{figure}[t]
\centering
\includegraphics[width=0.5\textwidth]{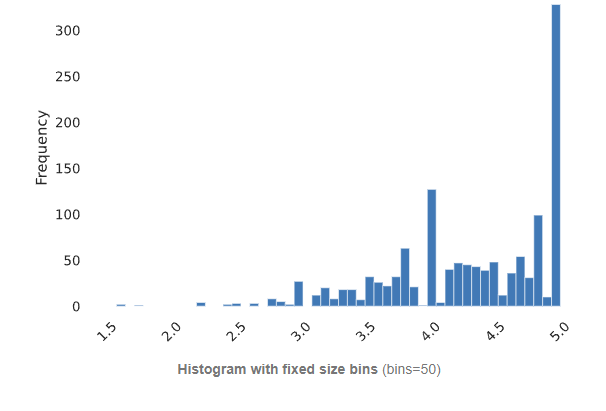}
\caption{Distribution of the user rating scores.
}
\label{fig:m}
\end{figure}
\subsection{Distribution of Dual-level Feedback Analysis}
\label{dis-feedback}
\noindent
\textbf{Stress Change}
Fig. \ref{fig:dis} (a) demonstrates the distribution of stress change for all conversations in the training set. Following the paper \cite{DBLP:conf/acl/LiuZDSLYJH20}, the help-seekers are prompted to fill in their stress levels before and after the supporting conversation. Then, this study uses the post-conversation stress level minus the prior-conversation stress level to obtain the conversation-level stress change, which reflects the users’ global states and thus can guide strategy selection from a global perspective. As shown in Fig. \ref{fig:dis} (a), the stress change of more than 88.46\% samples is less than zero (i.e., the user's emotional anxiety has been alleviated), indicating that the emotional stress of help-seekers can be relieved by utilizing the supportive strategies.

\noindent
\textbf{Relevance and Empathy Score}
During the post-conversation survey, the help-seekers are also asked to give a score, from 1 to 5, for the relevance and empathy level that the supporters provide during the conversations, where the value of -1 indicates that the help-seeker does not complete the survey. In the training process, the values of -1 are replaced with the average value is.  The distributions of empathy scores and relevance scores are shown in Fig. \ref{fig:dis} (b) and Fig. \ref{fig:dis} (c), respectively. It can be seen that about 81.69\% samples are located in 4 and 5 (good scores), which indicates that appropriate strategies are beneficial to providing supportive responses.

\noindent
\textbf{User Rating Score}
{Fig. \ref{fig:m}} shows the distribution of the user rating scores from the datasets. Following the paper \cite{DBLP:conf/acl/LiuZDSLYJH20}, the help-seekers are asked to give feedback after every two utterances they received from the supporters during the conversation. The rating scores are on a 5-star scale, and a higher score indicates better feedback. During the training, the user rating change will be calculated by the changes between the post-turn and prior-turn.

\begin{table}[t]
\centering
\setlength\tabcolsep{10pt}
\resizebox{\linewidth}{!}{
	\begin{tabular}{lcccc}
	\toprule
	\textbf{Strategy}& \textbf{B-2}\,$\uparrow$ &\textbf{B-4}\,$\uparrow$ & \textbf{D-1}\,$\uparrow$&\textbf{R-L}\,$\uparrow$\\ 
	\midrule
	Token& 6.12&{1.80}&{3.75}&{16.40}\\
	Description& \textbf{8.00}&\textbf{2.32}&\textbf{3.84}&\textbf{17.53}\\
	\toprule
	\end{tabular}}
\caption{Comparison of strategies at different granularities, including token level and description level.}
\label{tab:dictionary}
\end{table}
\subsection{Strategy Dictionary Analysis}
To quantitatively evaluate the benefit of the strategy dictionary, this study performs a comparison between the token-based and description-based information on strategy. The token-based method {follows the} paper \cite{DBLP:conf/acl/LiuZDSLYJH20}. As shown in Table \ref{tab:dictionary}, there is a performance drop in the token-based method, especially in B-$n$ and R-L. This indicates that the enriched strategy semantic information has the potential to understand strategies, thus generating a more supportive and strategy-constrained response.
\begin{table*}[htbp]
\centering	
\resizebox{\linewidth}{!}{
	\begin{tabular}{|l|l|}
	\multicolumn{2}{c}{\textbf{Case One: Context}}\\
	\hline
	Help-seeker &\makecell[l]{I'm good, but I need another job to pay more bills.}\\
	\hline
	Supporter &\makecell[l]{It seems you are having a difficult time with your finances. Is that correct?} \\
	\hline
	Help-seeker & \makecell[l]{Yes somewhat. I can pay it off within 2 or 3 years without a second job. I'd like to \textbf{\color{red}pay it off faster} \\ though therapist.} \\
	\hline
	\multicolumn{2}{|c|}{\textbf{Response}}\\
	\hline
	{{Ground-Truth}} & \makecell[l]{\textbf{\color{blue}(\textit{Affirmation and Reassurance})} I'm sure it'll take dedication and hard work, but I bet you can do!}\\
	\hline
	Blender-Bot & \makecell[l]{\textbf{\color{cyan}(\textit{Restatement})} That sounds like a difficult situation to be in. Is there anything you can do to help? }\\
	\hline
	MISC & \makecell[l]{\textbf{\color{blue}(\textit{Question})} I'm sorry to hear that. Is there anything specific you'd like to talk about? }\\
	\hline
	\textbf{FADO}  &\makecell[l]{\textbf{\color{blue}(\textit{Affirmation and Reassurance})} it is great that you are able to \textbf{\color{red}pay it off quickly}. It is important to \\ remember that you can pay it back quickly, if you want.}\\
	\toprule
	
	\multicolumn{2}{c}{\textbf{Case Two: Context}}\\
	\hline
	Help-seeker &\makecell[l]{I'm ok kind of stressed.}\\
	\hline
	Supporter &\makecell[l]{Anything I can help with :) ?} \\
	\hline
	Help-seeker & \makecell[l]{feeling stressed is never fun :( .}\\
	\hline
	&\makecell[l]{......}\\
	\hline
	Help-seeker & \makecell[l]{I want to declutter my house but \textbf{\color{red}don't know where to start}.}\\
	\hline
	\multicolumn{2}{|c|}{\textbf{Response}}\\
	\hline
	{{Ground-Truth}} & \makecell[l]{\textbf{\color{blue}(\textit{Restatement})} I hear you, feeling overwhelmed and like you don't know where to start can leave \\ a person feeling stressd forsure!}\\
	\hline
	Blender-Bot & \makecell[l]{\textbf{\color{blue}(\textit{Self-disclosure})} I understand how you feel. I recently started to feel overwhelmed as well. }\\
	\hline
	MISC & \makecell[l]{\textbf{\color{cyan}(\textit{Information})} I know how anxious and anxious you are right now. Do you have any idea how you \\ are going to make sure you have everything done correctly? }\\
	\hline
	\textbf{FADO}  &\makecell[l]{\textbf{\color{blue}(\textit{Self-disclosure})} I can understand that! It can be overwhelming. You are not alone in feeling over- \\whelmed. I have been struggling with finding a \textbf{\color{red}good place to start}, too.}\\
	\toprule
	
	\multicolumn{2}{c}{\textbf{Case Three: Context}}\\
	\hline
	Help-seeker &\makecell[l]{Just been having a rough time... I recently broke up with my girlfriend.}\\
	\hline
	Supporter &\makecell[l]{Sounds hard. Are you doing anything to help? } \\
	\hline
	Help-seeker & \makecell[l]{Trying to avoid thoughts about her, but every where on social media she is posting about \textbf{\color{red} her new} \\ \textbf{\color{red} guy}. It is hard to \textbf{\color{red}escape the sadness}.} \\
	\hline
	
	\multicolumn{2}{|c|}{\textbf{Response}}\\
	\hline
	{{Ground-Truth}} & \makecell[l]{\textbf{\color{blue}(\textit{Question})} Wow. In our connected world it is hard to get away. Do you want to tell me how it \\ happened?}\\
	\hline
	Blender-Bot & \makecell[l]{\textbf{\color{cyan}(\textit{Reflection of Feelings})} Are you doing anything to avoid thoughts about her? \textit{(Repetition)} }\\
	\hline
	MISC & \makecell[l]{\textbf{\color{cyan}(\textit{Providing Suggestions})} It sounds like she may be trying to focus on you as much as possible. Is \\ there any way you can get away from her? \textit{(Repetition)}}\\
	\hline
	\textbf{FADO}  &\makecell[l]{\textbf{\color{blue}(\textit{Reflection of Feelings})} I understand it is sad to \textbf{\color{red}see her new guy}. Have you tried to talk to her \\ about it?}\\
	\toprule
	
	\multicolumn{2}{c}{\textbf{Case Four: Context}}\\
	\hline
	Help-seeker &\makecell[l]{How did you come over that? I really need some guidance here. What do I do?}\\
	\hline
	Supporter &\makecell[l]{It came to me like a mood swing for so many days but I started going to meet friends and get cl-\\ oser to my friends for positive vibes?} \\
	\hline
	Help-seeker & \makecell[l]{You met with your friends? I've tried meeting with my friends. I feel ok when I'm with them, but \\ then later, I feel sad again. I can't spend all my time with my friends, of course. They have things \\ to do. So do I.} \\
	\hline
	\multicolumn{2}{|c|}{\textbf{Response}}\\
	\hline
	{{Ground-Truth}} & \makecell[l]{\textbf{\color{blue}(\textit{Question})} Yes, you have to be positive at all time and always do something that makes you happy?}\\
	\hline
	Blender-Bot & \makecell[l]{\textbf{\color{cyan}(\textit{Other})} I'm sorry to hear that. Is there anything you can do to help? }\\
	\hline
	MISC & \makecell[l]{\textbf{\color{cyan}(\textit{Restatement})} Sometimes it helps to start with small steps and then take them one by one. It's a \\ great way to start. }\\
	\hline
	\textbf{FADO}  &\makecell[l]{\textbf{\color{cyan}(\textit{Question})} I'm sorry to hear that. It sounds like you are not feeling well enough.}\\
	\hline
	\end{tabular}}
\caption{Case study of the generated responses by baselines and FADO. {\color{cyan}Cyan} indicates the inconsistency between the strategy and response, {\color{blue}blue} means consistency, and {\color{red}Red} represents the relevant information.}
\label{tab:generated}       
\end{table*}

\subsection{More Cases between FADO and the Baselines}
\label{morecase}
Table \ref{tab:generated} shows more examples to compare the FADO and the baselines qualitatively. For the first three cases, our FADO predicts the appropriate strategy and generates consistent responses (as shown in {\color{blue}blue} in Table \ref{tab:generated}) to comfort the help-seeker and provide some suggestions. By contrast, the baselines usually make an inconsistency between the strategy and response (as shown in {\color{cyan}cyan} in Table \ref{tab:generated}). Additionally, FADO produces more supportive responses that are relevant to the context (as shown in {\color{red}red} in Table \ref{tab:generated}), which illustrates the effectiveness of the proposed model. However, as for case four, the baselines and the FADO output an inconsistent response with the strategy. This may be because the model cannot well resolve the problem of long-distance dependence (the context is relatively long in the fourth case).

\subsection{Efficiency Analysis}
{To compare the efficiency between our model and other SOTA models, an extra experiment is carried out, as depicted in Table \ref{tab:latency}. This study calculates the running time of the model during the predicting phase { under the same setting for a fair comparison, e.g., the beam size is set to 1, and the validation batch size is set to 16. The settings of the Top-$p$ sampling rate and temperature are illustrated in Section \ref{setting}. These models are tested on the test dataset of ESConv \cite{DBLP:conf/acl/LiuZDSLYJH20}.} Compared with the previous methods, our model achieves an improvement in latency (\textbf{1.03} sentences per second). By contrast, GLHG performs the worst, and MISC reaches 0.94 sentences per second. {As for the reason of this phenomenon, GLHG constructs a hierarchical and complex graph that contains four types of nodes and three types of edges, resulting in a slower decoding speed. MISC considers attention mechanisms from multiple perspectives (e.g., mental states, strategy, and dialog history), so a large number of calculations affect the predictive efficiency of MISC. Without complex graph modeling and attention mechanism, FADO decodes faster than the SOTA models.} To sum up, {as listed in Table \ref{tab:main-exp} and Table \ref{tab:latency}}, the proposed model performs effectively and efficiently for the ESConv task.}

\subsection{Hyperparameter $\mu$ Analysis}
{To investigate how the conversation-level feedback affects the effectiveness of the FADO, this study manually sets the hyper-parameter $\mu$ (Eqn. \ref{eq:score}). As shown in Table \ref{tab:mux}, the performance of the model reaches 16.98\% on ROUGE-L when removing the conversation-level feedback ($\mu$ = $0$), but the model achieves a decent result when $\mu$ is set to $0.5$. When more conversation-level feedback is incorporated, the performance of the FADO tends to be degressive.}

\begin{table}[h]
\centering
\resizebox{0.85\linewidth}{!}{
	\begin{tabular}{lcc}
	\toprule
	\textbf{Model}	& \textbf{Latency (sentences/s)} & $\Delta$  \\ \midrule
	~{GLHG} 	& {0.81}	&  -
	\\
	~{MISC} & {0.94}	& + 0.13	\\
	~{FADO} (Ours)	& \textbf{1.03}	&	+ 0.22	\\
	\toprule
	\end{tabular}}
\caption{\label{tab:latency} The results of efficiency analysis between the proposed model and new baselines. }
\end{table}

\section{Conclusion}
In this paper, a {F}eedback-{A}ware {D}ouble C{o}ntrolling Network is proposed to make a strategy schedule and generate supportive responses. Different from previous studies, the DFS leverages dual-level feedback to encourage or penalize strategies rather than simply optimizing with the ground truth. Meanwhile, the DCR models {C2S} and {S2C} flows to generate strategy-constrained responses, which improves the consistency score significantly. 
Quantitative results on the ESConv indicate that the proposed model achieves the SOTA performance. Also, qualitative analyses demonstrate the importance of each component in FADO. For future work, some other attributions of help-seekers will be considered, such as personal information, educational background, etc., which need further investigation in emotional support scenarios.
\begin{table}[b]
\centering
\begin{tabular}{lcccc}
\toprule
\textbf{Method}& \textbf{ACC}\,$\uparrow$ &\textbf{B-2}\,$\uparrow$ & \textbf{D-1}\,$\uparrow$&\textbf{R-L}\,$\uparrow$\\ 
\midrule
BlenderBot& 28.11&{5.83}&{2.60}&{15.57}\\
\midrule
FADO($\mu$=0)         & {30.96}	& {7.84}  & {3.66}  & {16.98}      \\
{FADO($\mu$=0.2)} & {32.13} & {7.64} & {3.62} & {17.45} \\
\textbf{FADO($\mu$=0.5)  }   & \textbf{32.90} & \textbf{8.00} & \textbf{3.84} & \textbf{17.53}      \\
FADO($\mu$=0.8)          & 31.02          & 7.68          & 3.62          & 17.40          \\
FADO($\mu$=1.0)            & 31.92          & 7.75          & 3.60          & 17.41   \\
\toprule
\end{tabular}
\caption{\label{tab:mux} Results of different setting on the hyper-parameters $\mu$.}
\end{table}
\section*{Acknowledgments}
We thank all anonymous reviewers for their constructive comments. This work is supported by the National Natural Science Foundation of China (No.U21B2009).



\printcredits

\bibliographystyle{cas-model2-names}

\bibliography{cas-refs}


\bio{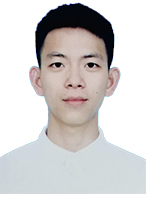}
{Wei Peng} received the B.S. degree in Computer Science and Technology from Chang'an University, Xi'an, China, in 2018. And now he is currently in the stage of M.S. and Ph.D in Institute of Information Engineering, Chinese Academy of Sciences. He has authored or coauthored about 20 papers in question answering and dialog system area. His research interests include sentiment analysis, empathetic dialog generation and question answering.
\endbio

\newpage
\bio{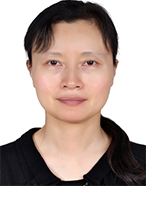}
{Yue Hu} received the Ph.D. degree from University of Science and Technology Beijing, Beijing, China, in 2000. She is currently a Professor with the School of Cyber Security, University of Chinese Academy of Science, Beijing, China. She has authored or coauthored about 100 papers in various journals and proceedings. Her research interests include machine translation, information extraction, question answering, and dialog system.
\endbio

\bio{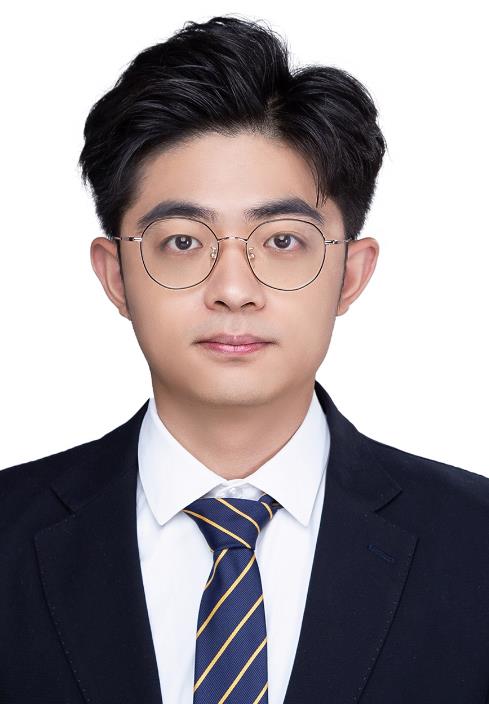}
{Yuqiang Xie} received the B.S. degree in Network Engineering from China University of Mining and Technology, Xu'zhou, China, in 2017. And now it is currently in the stage of M.S. and Ph.D in Institute of Information Engineering, Chinese Academy of Sciences. His research interests include story understanding and generation, cognitive modeling, affective computing and question answering.
\endbio
\vspace{21pt}
\bio{}
Ziyuan~Qin received the B.S. degree in Mathematcis from Ohio State University, Columbus, United States, in 2016 and a M.S. degree in Information Technology from the University of New South Wales, Sydney, Australia, in 2018. His research interests include multi-modal sentiment analysis, vision-language learning, and visual-question answering.
\endbio
\vspace{21pt}
\bio{}
{Yunpeng Li} received the B.S. degree in Computer Science and Technology from Shandong University, Jinan, China, in 2019. And now it is currently in the stage of M.S. and Ph.D in Institute of Information Engineering, Chinese Academy of Sciences. His research interests include dialog generation, and question answering.
\endbio

\end{document}